\documentclass[10pt,twocolumn,letterpaper]{article}

\usepackage{cvpr}
\usepackage{times}
\usepackage{epsfig}
\usepackage{graphicx}
\usepackage{amsmath}
\usepackage{amssymb}
\usepackage{multirow}
\usepackage{enumitem}

\usepackage[breaklinks=true,bookmarks=false]{hyperref}

\cvprfinalcopy 

\def\cvprPaperID{0008} 

\begin{document}

\title{YUVMultiNet: Real-time YUV multi-task CNN for  autonomous driving}

\author{Thomas Boulay$^{1}$, Said El-Hachimi$^{1}$, Mani Kumar Surisetti$^{2}$, Pullarao Maddu$^{3}$, Saranya Kandan$^{2}$\\
$^{1}$Valeo France, $^{2}$Valeo India, $^{3}$Valeo Ireland\\
{\tt\small \{thomas.boulay, said.el-hachimi, manikumar.surisetti, pullarao.mbv, saranya.kandan\}@valeo.com}
}

\maketitle

\begin{abstract}
   In this paper, we propose a multi-task convolutional neural network (CNN) architecture optimized for a low power automotive grade SoC. We introduce a network based on a unified architecture where the encoder is shared among the two tasks namely detection and segmentation. The proposed network runs at $25$ FPS for $1280\times800$ resolution. We briefly discuss the methods used to optimize the network architecture such as using native YUV image directly, optimization of layers \& feature maps and applying quantization. We also focus on memory bandwidth in our design as convolutions are data intensives and most SOCs are bandwidth bottlenecked.
   We then demonstrate the efficiency of our proposed network for a dedicated CNN accelerators presenting the key performance indicators (KPI) for the detection and segmentation tasks obtained from the hardware execution and the corresponding run-time.  
\end{abstract}

\section{Introduction}

Current improvements in the field of Advanced driver-assistance systems (ADAS) are going to confirm that Deep-Learning become essential to propose efficient solutions based on camera sensors. Various visual perception tasks including semantic segmentation, bounding box object detection or depth estimation are nowadays successfully handled by Convolutional Neural Networks (CNNs). However, most of these successful applications generally needs a high computational power which is not realistic for the current embedded systems. In this paper, we propose a new real-time multi-task network for object detection and semantic segmentation. Our paper propose two major improvements in comparison with the recent solutions. First we propose a multi-task network fed by YUV$4$:$2$:$0$ images. Secondly, we propose a network fully designed in order to take all the key factors for an efficient embedded integration into consideration. The paper is structured as follows. The section \ref{sec:PriorWork} is a review of the recent CNN applications in various domain and for different types of image colorspace (RGB or YUV). The section \ref{sec:hardware} is a high-level description of the dedicated CNN hardware used to integrate our proposed network. The section \ref{sec:NetworkOptimization} describes in detail the network design and the optimization done for an efficient integration on the selected low power SoC. In the section \ref{sec:Quantization}, some details are given on how to design our network to take into consideration this quantization stage. Finally, some results are presented in the section \ref{sec:Results} and details are given about our demo proposal.

\begin{figure}[t]
\begin{center}
   \includegraphics[width=0.8\linewidth]{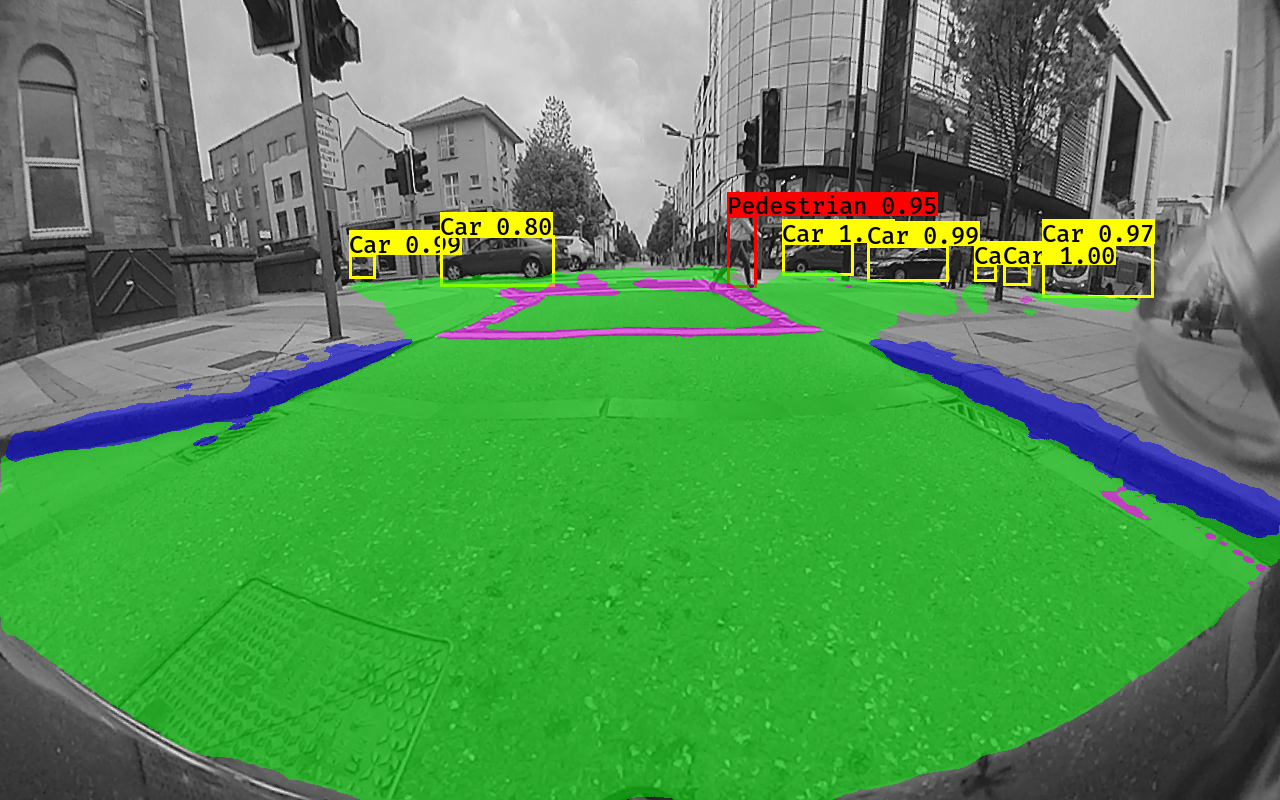}
\end{center}
   \caption{Example of forward pass. Vehicle/Pedestrian detection and road/lanes/curb segmentation}
\label{fig:mtl_output}
\end{figure}

\section{Background}

\textbf{Related Work:}
\label{sec:PriorWork}
For several years, the architecture of the CNN was dedicated to a single task like the classification task \cite{NIPS2012_4824}, \cite{SimonyanZ14a}, \cite{HeZRS15}, detection task \cite{Yolov3}, \cite{SSD}), semantic segmentation task \cite{PSPNet}, \cite{RefineNet}, \cite{DeepLabv3extend} or visual SLAM \cite{visualSlam}. In the recent year, we observe the appearance of new CNN architectures designed specifically for handling several tasks. Among these new approaches, Teichmann \cite{TeichmannWZCU16} in 2016 proposed an unified encoder-decoder architecture which can perform three tasks (classification, detection, segmentation) simultaneously. In 2018, \cite{GansehIsa2019} presented a Multinet-based paper lightened to achieve 30 fps on a low-power embedded system. In addition to these papers, several others \cite{Fukuda2018}, \cite{Bischke2017} confirmed the benefit of using multi-task learning (MTL) method especially for computational efficiency concerns or for assuring a best generalization and accuracy of the network. Our work is inspired by these papers but we propose a network even more lightened, using directly YUV$4$:$2$:$0$ input images to reduce the bandwidth and a demo is presented from this work on low-power SoC. 
In 2013, Sermanet and LeCun \cite{Sermanet2012} proposed to use YUV image space for their pedestrian detection application. The input layer sequence of our network is directly inspired from their work except that we propose an adaption of the kernel dimensions ($5 \times 5$ and $3 \times 3$ instead of $7 \times 7$ and $5 \times 5$) and some other details for fitting our SoC constraints. \\

\textbf{The dedicated CNN hardware:} \label{sec:hardware}
The dedicated CNN hardware we have chosen for our experiment is a low power automotive SOC. It is able to execute around $400$ Giga Multiplication Accumulation per second (MACS). In one cycle, the CNN IP processes $4$ input channels, each of them is convolved by $8$ convolution kernels to generate the $8$ output channels. $100 \%$ CNN core utilization will be achieved by using $5 \times 5$ convolution kernels. The weights and the input/output channels are encoded as 16-bit fixed-point numbers.

\begin{figure}[t]
\begin{center}
   \includegraphics[width=0.9\linewidth]{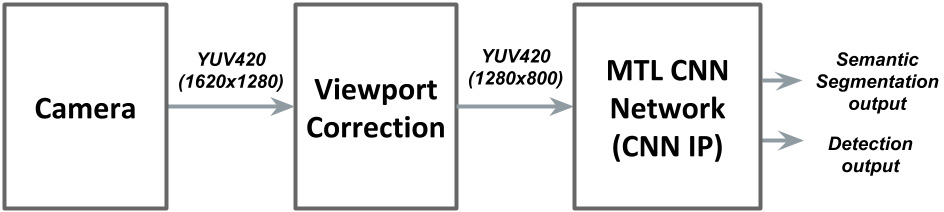}
\end{center}
   \caption{SoC video pipeline}
\label{fig:pipeline}
\end{figure}

The CNN unit is a part of a SoC containing several dedicated units. The SoC is designed to handle an automotive real-time application from the image acquisition through common camera connectivity to the final display. The real-time video pipeline used for our application is described in figure \ref{fig:pipeline}.

\section{Network architecture optimization}
\label{sec:NetworkOptimization}
In this section, we describe network optimization stages. The three high level stages of optimization blocks are: Input data shape optimization, Encoder optimization and Decoders optimization.

The optimizations are targeted to achieve the peak processing power of the CNN hardware accelerator in context. These optimization steps bring our base network to major changes that we will describe in the following section.

\subsection{Input data shape optimization}
To meet the specific data format provided by the selected SoC, the network input layers had to be modified. Indeed, the selected SoC is optimized for its application in front camera. In our case, the camera used are fisheye cameras. This type of camera offers a much wider field of view than the fixed camera, which is useful for automotive applications. The fisheye correction is done by an image render for camera image distortion correction that presented on SOC and the corrected image is sent to the CNN dedicated IP. The image format delivered by the Image Signal Processor (ISP) at the beginning of the video pipeline isn't BGR format but rather a YUV format. This type of format is relatively common for the embedded systems but this type of format is not often used for the publication in the deep-learning field. The YUV format selected for our application is a YUV$4$:$2$:$0$ format. Compared to a common BGR image, the three channels that encode the information captured by the camera are not at the same resolution. The channel U and V representing the chrominance components are half resolution than the channel Y representing the luma. That means that one pixel in the channel U or V encodes the chrominance information corresponding to a $2 \times 2$ block of pixels in the Y channel.

\begin{figure}[t]
\begin{center}
   \includegraphics[width=0.9\linewidth]{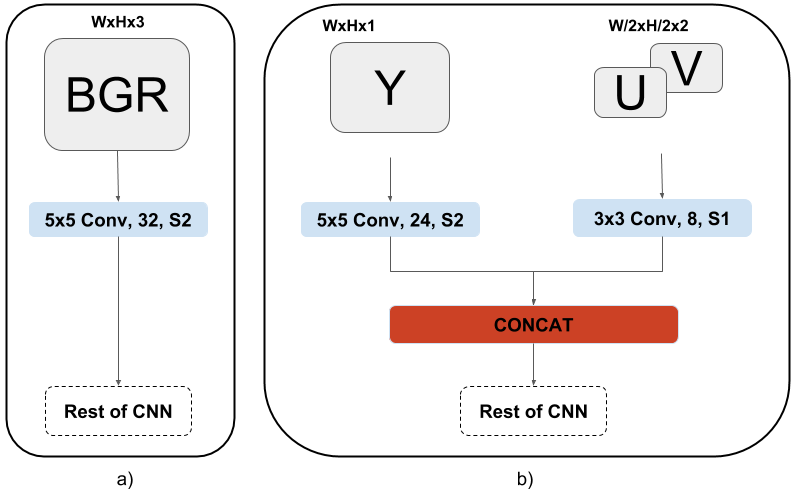}
\end{center}
   \caption{a) Common BGR input data layer, b) our optimized YUV$4$:$2$:$0$ data layer (right)}
\label{fig:YUV420_input_data_shape}
\end{figure}

The two options presented in the figure \ref{fig:YUV420_input_data_shape} have exactly the same cost on the CNN IP. However, the YUV$4$:$2$:$0$ option has two major advantages: the memory bandwidth reduction by factor $2$ and there is no need for a additional conversion (YUV$4$:$2$:$0$ to BGR) module.

Let's assume a $1280 \times 800$ 8-bit input image. For the BGR option the memory traffic to feed the network is around $3$ MBytes, for the YUV$4$:$2$:$0$ option, it's around $1.5$ MBytes. Another advantage of the YUV$4$:$2$:$0$ option is there is no need for an additional module for YUV$4$:$2$:$0$ to BGR conversion. Using the BGR option, this would have been mandatory to give something valid as input of the CNN.

\subsection{Encoder stage : V2N9Slim}
\begin{figure}[t]
\begin{center}
   \includegraphics[width=0.9\linewidth]{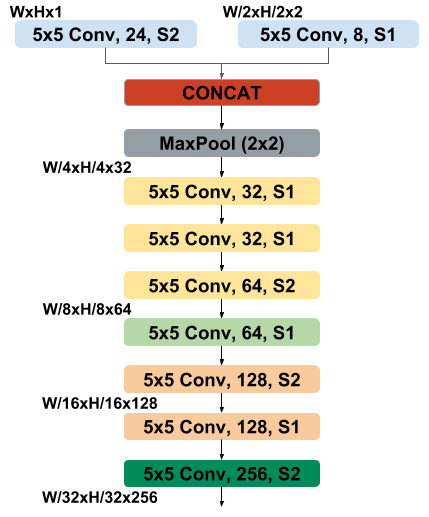}
\end{center}
   \caption{V2N9 encoder architecture}
\label{fig:V2N9_arch}
\end{figure}

In the most common unified encoder-decoder architectures, the most expensive part is generally the encoder part. 
VGG \cite{SimonyanZ14a} or ResNet \cite{HeZRS15} based network are basically the most common encoders that we can find in the recent papers.  
Unfortunately, these encoders are wide and deep and aren't de facto good candidates for low power embedded system. 
To meet the embedded system constraints, a new light-encoder called V2N9Slim is employed.
V2N9Slim is a $9$ layers encoder ($9$ Convolution or Pooling layers). The encoder architecture is described in the figure \ref{fig:V2N9_arch}. The V2N9Slim architecture is a mixture of Resnet and VGG architectures. Indeed, the layer sequence of our network is mainly inspired by the plain network of Resnet, no shortcut connections are used for our network which come up to the VGG architecture. The motivation for not using the shortcut connections comes from previous experiments which are detailed in this paper \cite{Das2018}. The authors demonstrate that the shortcut connections don't provide a significant improvement in accuracy in shallow networks, which is the case with our $9$ layers encoder. Furthermore, the shortcut connections are also costly for the embedded systems since they involves additional computation and an increase of the memory footprint and the memory transfers.

For the multi-task experiments, the encoder part is pre-trained on a classification task using the ImageNet dataset \cite{ILSVRC15}.

\subsection{Convolution shape}

The $9$ convolutions of V2N9Slim network are $5 \times 5$ convolution. As we already said, in \ref{sec:hardware}, the hard-wired operations on the CNN IP of our low power automotive SoC are $5 \times 5$ convolution. Convolution $3 \times 3$ or $1 \times 1$ are possible on the SoC but the MAC resources are inefficiently used (only $36 \%$ or $4 \%$ of multipliers used respectively). $7x7$ convolution are also possible on the SoC but once again in term of efficiency it involves for each convolution $1$ cycle more than a convolution with a kernel size equal to $5 \times 5$.
Several variants of the V2N9Slim encoder have been tested for the classification task on the ImageNet challenge. We tested with a $7\times7$ kernel on the first convolution or replacing the $5\times5$ kernels by $3\times3$ kernels on all the layers except the first one. All these experiments are given classification accuracies very similar but the variant described on the figure \ref{fig:V2N9_arch} slightly over-performed the other ones.

%
%

\subsection{Layer sequence}

Each convolution layer presented on the figure \ref{fig:V2N9_arch} are followed by a Batchnorm layer and a ReLU layer. It doesn't appear on the figure for the sake of readability. These two layers are free on the SoC (under some conditions) just like the Pooling, Concatenate or Layer-Wise addition layers. The condition is that all these layers must follow a convolution layer. Indeed, the CNN core on the SoC is designed to execute a sequence of layers in one clock cycle and this sequence always starts by a convolution layer. 
We can find for example in many papers, a Batchnorm layer directly on the input data to normalize and center the data distribution. Such operations are definitively not possible on the SoC because the batch normalization operation is not directly hard-wired on the silicon. A solution is to fuse the batchnorm parameters into the first convolution weights and bias. Indeed the scale factor can be apply on the convolution weights and the offset parameter in the bias. However, it is not possible for some specific cases like a convolution with zero padding because the bias is added on the zero-padded feature maps. It introduces artefacts on the border of the feature maps during the inference execution since the training phase didn't take into consideration this specificity. A solution to prevent these type of issues is to adapt the training to our specific CNN core that means for instance train our network with a constant-pad layer instead of a zero-pad layer to handle the offset parameter. 

\subsection{Decoders}
The decoders for our proposed solution are YoloV2 decoder for the detection part and a FCN based decoder for the segmentation part. The YoloV2 decoder consist to add one convolutional layer at the end of the encoder to reshape the encoder feature maps to bounding-box predictions. Then, an additional post-processing is executed on the ARM processor to transform the output of the YoloV2 decoder into bounding boxes coordinates and confidences. The FCN based decoder is a decoder with 5 transposed convolution layer stages. Each of these layers up-samples (using a $5 \times 5$ kernel) the input by factor 2 (ie FCN$2$) for retrieving the original input image resolution. The FCN$2$ decoder is entirely run on the dedicated CNN IP present on the SoC.

\begin{table*}[]
\centering
\begin{tabular}{|l|c|l|l|l|l|c|l|l|l|l|}
\hline
 \multirow{3}{*}{\textbf{Network Model}} & \multicolumn{5}{c|}{\textbf{Segmentation}} & \multicolumn{3}{c|}{\textbf{Detection}} & \multicolumn{2}{l|}{\textbf{Runtime}} \\ \cline{2-11} 
 & \multirow{2}{*}{mIOU} & \multicolumn{4}{c|}{Class IoU} & \multirow{2}{*}{mAP} & \multicolumn{2}{c|}{Class AP} & \multicolumn{2}{l|}{} \\ \cline{3-6} \cline{8-11} 
 &  & Void & Road & Lane & Curb &  & Car & Ped & ms & fps \\ \hline
{Reference Model} & \multicolumn{1}{l|}{74.3} & 97.9 & 94.6 & 61.5 & 43.6 & \multicolumn{1}{l|}{34.7} & 54.5 & 14.9 & 66 & 15 \\ \hline
\textbf{Proposed Model} & \multicolumn{1}{l|}{\textbf{74.7}} & \textbf{98.0} & \textbf{94.6} & \textbf{61.7} & \textbf{44.4} & \multicolumn{1}{l|}{\textbf{47.6}} & \textbf{66.2} & \textbf{29.1} & \textbf{40} & \textbf{25} \\ \hline
\end{tabular}
\caption{Performance Comparison of Models}
\label{tab:Results}
\end{table*}

\subsection{Quantization steps}
\label{sec:Quantization}
Our proposed optimized architecture has been trained using a Keras based framework in floating-point precision. To execute the network trained in floating-point precision on the 16-bit fixed-point precision, a quantization step is required to run the network on the dedicated hardware.
To reduce as much as possible issues like saturation during this stage, all the convolutional layers are followed by a batch-normalization layer to have the feature maps distribution with an unit variance. It also makes more accurate the encoding of the floating point numbers into 16-bit numbers since the floating point range is limited by the batch-normalization.

\section{Results}
\label{sec:Results}

We performed our set of experiments on our private fisheye camera dataset. This dataset has a size of $5000$ annotated frames. The image resolution is $1280 \times 800$. The frames are annotated for both semantic segmentation and detection tasks. For semantic segmentation, each pixel on the frames is annotated to belong to one of the $4$ following classes : [Background, Road, Lane, Curb]. For detection class, the object in the frame are annotated among $2$ classes: [Pedestrian, Car].\\

\noindent Two types of network are evaluated:

\begin{enumerate} [nosep]
    \item \textbf{V2N9slim+YoloV2+FCN$2$ (Proposed Model): }
Our proposed network is composed by the V2N9Slim encoder and YoloV2, FCN$2$ for the decoders. This network is fed by YUV$4$:$2$:$0$ images.
    \item \textbf{Resnet10slim+YoloV2+FCN$2$ (Reference Model): }
A baseline network composed by a Resnet10Slim based encoder and the two same decoders. This network takes as input BGR images. \\
\end{enumerate}

On the table \ref{tab:Results}, the performances of these two networks are summarized. We can observe that the KPI on the segmentation task and the detection task are very close for the two networks. Our proposed network is able to achieve the same level of accuracy and even better than the Multinet reference network. However, our proposed network is $65\%$ faster than the reference network which proves the efficiency of the optimization stages presented in the section \ref{sec:NetworkOptimization}.


%

\section{Demo proposal}
The purpose of the demo is to show real-time execution of our network on the low power SoC. Basically two options are considered : run our network on a set of images pre-loaded on the hardware or run our network using the full pipeline, it means from the image acquisition with the camera to the network output display. 
For the two options, the outputs of the CNN prediction (segmentation and detection) have to be post-processed for displaying on the demo screen.

The demo video attached to this paper is the result of the first option execution following by the detection post-processing for the Bounding-Boxes display and by the segmentation post-processing for the labelled frames display. These two post-processing are executed on the ARM core present on the SoC and are executed sequentially after the CNN core execution. On our final application, these post-processing will be set up in parallel with CNN inference to improve execution run-time.

The first option is what we submitted but we are working on the second option and pipeline optimization to have a final version of our demonstration ready before the workshop.
Please find in the following link the demo video:
\href{https://drive.google.com/open?id=1cVLxRaqwa9iwHcKkHB0CQiwsIGeeOGGY}{Demo video}

\section{Conclusion}
In this paper, we presented an optimized multi-task network for automated driving application. First, we described briefly our low power SoC and motivated the need to use the multi-task approach. Then, we described the proposed network and the optimizations done to achieve the embedded constraints. Finally, we shared experimental results on our own fisheye dataset and on the low power SoC.
We can reasonably maintain that three main contributions are presented in this paper: an input data layer dedicated and optimized for YUV$4$:$2$:$0$ image space, a network oriented towards the CNN core of a low power SoC and a network highly reduced achieving accuracy close to the original Multinet network but much faster. The demo proposal for the workshop is a good opportunity to highlight these contributions and show live the efficiency of our network.

\newpage

{\small
\bibliographystyle{ieee}
\bibliography{valeoCVPR}
}

\end{document}